\documentclass[11pt,letterpaper]{article}
\usepackage{acl2017}
\usepackage{times}
\usepackage{latexsym}
\usepackage{amsmath, amssymb}
\usepackage{multirow}
\usepackage{xspace}

\usepackage{graphics}
\usepackage[nomessages]{fp}

\def\roundposition{1}

\edef\rounded{0}
\newcommand{\rdm}[1]{\edef\rounded{0}\FPeval\rounded{round(#1,\roundposition)}\rounded}

\graphicspath{ {fig/}{} }

\aclfinalcopy % Uncomment this line for the final submission

\newcommand\BLEU{\textsc{Bleu}\xspace}
\newcommand\TER{\textsc{Ter}\xspace}

\newenvironment{packed_descr}{
\begin{description}
  \setlength{\itemsep}{1pt}
  \setlength{\parskip}{-1pt}
  \setlength{\parsep}{0pt}
}{\end{description}}

\title{Beam Search Strategies for Neural Machine Translation}

\author{Markus Freitag \and Yaser Al-Onaizan\\
           IBM T.J. Watson Research Center\\
        1101 Kitchawan Rd, Yorktown Heights, NY 10598\\
        \{freitagm,onaizan\}@us.ibm.com}

\date{}

\begin{document}

\maketitle

\begin{abstract}
The basic concept in Neural Machine Translation (NMT) is to train a large Neural Network that maximizes the translation performance on a given parallel corpus. NMT is then using a simple left-to-right beam-search decoder to generate new translations that approximately maximize the trained conditional probability.
The current beam search strategy generates the target sentence word by word from left-to-right while keeping a fixed amount of active candidates at each time step. 
First, this simple search is less adaptive as it also expands candidates whose scores are much worse than the current best.
Secondly, it does not expand hypotheses if they are not within the best scoring candidates, even if their scores are close to the best one.
The latter one can be avoided by increasing the beam size until no performance improvement can be observed. 
While you can reach better performance, this has the drawback of a slower decoding speed.
In this paper, we concentrate on speeding up the decoder by applying a more flexible beam search strategy
whose candidate size may vary at each time step depending on the candidate scores.
We speed up the original decoder by up to 43\% for the two language pairs German$\to$English and Chinese$\to$English without losing any translation quality.
\end{abstract}

\section{Introduction}
\label{sec:introduction}
Due to the fact that Neural Machine Translation (NMT) is reaching comparable or even better performance compared to the traditional statistical machine translation (SMT) models \cite{jean+:2015,luong+:2015}, it has become very popular in the recent years~\cite{kalchbrenner+blunsom:2013,sutskever+:2014,bahdanau+:2014}. 
With the recent success of NMT, attention has shifted towards making it more practical. One of the challenges is the search strategy for extracting the best translation for a given source sentence. 
In NMT, new sentences are translated by a simple beam search decoder that finds a translation that approximately maximizes the conditional probability of a trained NMT model. The beam search strategy generates the translation word by word from left-to-right while keeping a fixed number (beam) of active candidates at each time step. By increasing the beam size, the translation performance can increase at the expense of significantly reducing the decoder speed.
Typically, there is a saturation point at which the translation quality does not improve any more by further increasing the beam.
The motivation of this work is two folded. First, we prune the search graph, thus, speed up the decoding process without losing any translation quality. 
Secondly, we observed that the best scoring candidates often share the same history and often come from the same partial hypothesis. We limit the amount of candidates coming from the same partial hypothesis to introduce more diversity without reducing the decoding speed by just using a higher beam.

\section{Related Work}
\label{sec:related_work}

The original beam search for sequence to sequence models has been introduced and described by \cite{graves2012sequence,boulanger2013audio} and by \cite{sutskever+:2014} for neural machine translation. 
\cite{hu2015improved,mi2016vocabulary} improved the beam search with a constraint softmax function which only
considered a limited word set of translation candidates to reduce the computation complexity. This has the advantage that they normalize only a small set of
candidates and thus improve the decoding speed.
\cite{wu2016google} only consider tokens that have local scores that are not more than beamsize below the best token during their search.
Further, the authors prune all partial hypotheses whose score are beamsize lower than the best final hypothesis (if one has already been generated).
In this work, we investigate different absolute and relative pruning schemes which have successfully been applied in statistical machine translation
for e.g. phrase table pruning \cite{zens2012systematic}.

\section{Original Beam Search}
The original beam-search strategy finds
a translation that approximately maximizes the conditional probability given by
a specific model. It builds the translation from left-to-right and
keeps a fixed number (beam) of translation candidates with the highest log-probability at each time step.
For each end-of-sequence symbol that is selected among the
highest scoring candidates the beam is reduced by one and the translation is 
stored into a final candidate list.
When the beam is zero, it stops the search and picks the translation with the highest
log-probability (normalized by the number of target words) out of the final candidate list.

\section{Search Strategies}
\label{sec:search_strategies}

In this section, we describe the different strategies we experimented with. 
In all our extensions, we first reduce the candidate list to the current beam size and apply on top of this 
one or several of the following pruning schemes.

\begin{packed_descr}

\item[Relative Threshold Pruning.]
The relative threshold pruning method discards
those candidates that are far worse than the best active candidate.
Given a pruning threshold $rp$ and an active candidate list $C$,
a candidate $cand \in C$ is discarded if:
\begin{equation}
score(cand) \leq rp * \max\limits_{c \in C} \{score(c)\}
\end{equation}

\item[Absolute Threshold Pruning.]
Instead of taking the relative difference of the scores into account, we just discard
those candidates that are worse by a specific threshold than the best active candidate.
Given a pruning threshold $ap$ and an active candidate list $C$,
a candidate $cand \in C$ is discarded if:
\begin{equation}
score(cand) \leq \max\limits_{c \in C} \{score(c)\} - ap
\end{equation}

\item[Relative Local Threshold Pruning.]
In this pruning approach, we only consider the score $score_w$ of the last generated word and
not the total score which also include the scores of the previously generated words.
Given a pruning threshold $rpl$ and an active candidate list $C$,
a candidate $cand \in C$ is discarded if:
\begin{equation}
score_w(cand) \leq rpl * \max\limits_{c \in C} \{score_w(c)\}
\end{equation}

\item[Maximum Candidates per Node]
We observed that at each time step during the decoding process, most of the partial hypotheses
share the same predecessor words. To introduce more diversity,
we allow only a fixed number of candidates with the same history at each time step.
Given a maximum candidate threshold $mc$ and an active candidate list $C$,
a candidate $cand \in C$ is discarded if already $mc$ better scoring partial hyps with the same history are in the candidate list.

\end{packed_descr}

\section{Experiments}
\label{sec:experiments}
For the German$\rightarrow$English translation task, we train an NMT system based on the WMT 2016 training data \cite{bojar2016findings} (3.9M parallel sentences).
For the Chinese$\rightarrow$English experiments, we use an NMT system trained on 11 million sentences from the BOLT project. 

In all our experiments, we use our in-house attention-based NMT implementation which is similar to ~\cite{bahdanau+:2014}.
For German$\rightarrow$English, we use sub-word units extracted by byte pair encoding~\cite{sennrich2015neural} instead of words
which shrinks the vocabulary to 40k sub-word symbols for both source and target.
For Chinese$\rightarrow$English, we limit our vocabularies to
be the top 300K most frequent words for both source and target language.
Words not in these vocabularies are converted into an unknown token. 
During translation, we use the alignments (from the attention mechanism) to replace the unknown tokens either with
potential targets (obtained from an IBM Model-1 trained on the parallel data) or with the source word itself (if no target was found) \cite{mi2016vocabulary}. 
We use an embedding dimension of 620 and fix the RNN GRU layers to be of 1000 cells each. For
the training procedure, we use SGD~\cite{bishop1995neural} to update model parameters with a mini-batch size of 64.
The training data is shuffled after each epoch.

We measure the decoding speed by two numbers. First, we compare the actual speed relative to the same setup without any pruning. Secondly, we
measure the average fan out per time step. For each time step, the fan out is defined as the number of candidates we expand. Fan out has
an upper bound of the size of the beam, but can be decreased either due to early stopping (we reduce the beam every time we predict a end-of-sentence symbol) or by the proposed pruning schemes.
For each pruning technique, we run the experiments with different pruning thresholds and chose the largest threshold that did not degrade
the translation performance based on a selection set.

\begin{table*}[t!]
\begin{center}
    {\setlength{\tabcolsep}{.3em}
    \begin{tabular}{|l|c|c|c|c|cc|cc|}
        \hline
        \bf{pruning}& \bf{beam} & \bf{speed} & \bf{avg fan out} & \bf{tot fan out} & \multicolumn{2}{c}{\bf{newstest2014}} & \multicolumn{2}{c|}{\bf{newstest2015}} \\
                    & \bf{size} & \bf{up} & \bf{per sent} & \bf{per sent} & \BLEU & \TER  & \BLEU & \TER \\ \hline \hline
        no pruning & 1 & - & 1.00 & 25 & \rdm{25.51}& \rdm{56.81}& \rdm{26.12}& \rdm{55.43}\\ \hline
        no pruning & 5 & - & 4.54 & 122 & \rdm{27.32}& \rdm{54.64}& \rdm{27.43}& \rdm{53.72}\\ 
        rp=0.6 & 5 & 6\% & 3.71 & 109 & \rdm{27.26}& \rdm{54.68}& \rdm{27.34}& \rdm{53.78}\\ 
        ap=2.5 & 5 & 5\% & 4.11 & 116 & \rdm{27.33}& \rdm{54.61}& \rdm{27.40}& \rdm{53.73}\\ 
        rpl=0.02 & 5 & 5\% & 4.25 & 118 & \rdm{27.29}& \rdm{54.69}& \rdm{27.38}& \rdm{53.77}\\ 
        mc=3 & 5 & 0\% & 4.54 & 126 & \rdm{27.42}& \rdm{54.57}& \rdm{27.49}& \rdm{53.75}\\ 
        rp=0.6,ap=2.5,rpl=0.02,mc=3 & 5 & 13\% & 3.64 & 101 & \rdm{27.26}& \rdm{54.61}& \rdm{27.32}& \rdm{53.77} \\ \hline
        no pruning & 14 & - & 12.19& 363 & \rdm{27.60}& \rdm{54.32}& \rdm{27.60}& \rdm{53.46}\\
        rp=0.3 & 14 & 10\% & 10.38& 315 & \rdm{27.59}& \rdm{54.28}& \rdm{27.60}& \rdm{53.44}\\ 
        ap=2.5 & 14 & 29\% & 9.49 & 279 & \rdm{27.55}& \rdm{54.33}& \rdm{27.60}& \rdm{53.52}\\
        rpl=0.3 & 14 & 24\% & 10.27& 306 & \rdm{27.56}& \rdm{54.37}& \rdm{27.67}& \rdm{53.42}\\ 
        mc=3 & 14 & 1\%  & 12.21& 347 & \rdm{27.63}& \rdm{54.38}& \rdm{27.66}& \rdm{53.38}\\
        rp=0.3,ap=2.5,rpl=0.3,mc=3 & 14 & 43\% & 8.44 & 260 & \rdm{27.59}& \rdm{54.45}& \rdm{27.64}& \rdm{53.43}\\ \hline
        rp=0.3,ap=2.5,rpl=0.3,mc=3 & - & - & 28.46 & 979 & \rdm{27.57}& \rdm{54.37}& \rdm{27.62}& \rdm{53.33}\\ \hline
\end{tabular}
\caption{Results German$\rightarrow$English: relative pruning(rp), absolute pruning(ap), relative local pruning(rpl) and maximum candidates per node(mc). Average fan out is the average number of candidates we keep at each time step during decoding.}
\label{tab:results-deen}
}
\end{center}
\end{table*}

\begin{figure}[t!]
\centering
\resizebox{\linewidth}{!}{\input{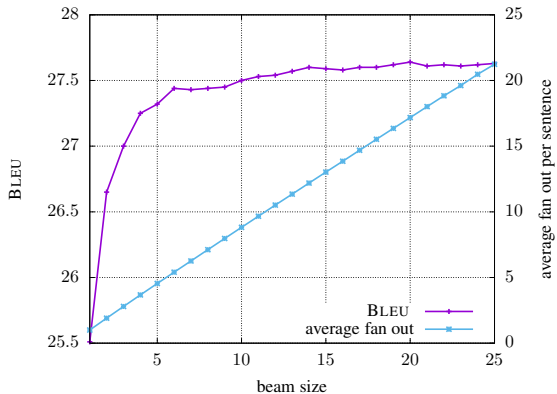}}
\vspace{-2em}
\caption{German$\rightarrow$English: Original beam-search strategy with different beam sizes on newstest2014.}
\label{fig:continue_ensemble}
\end{figure}

\begin{figure}[t!]
\centering
\resizebox{\linewidth}{!}{\input{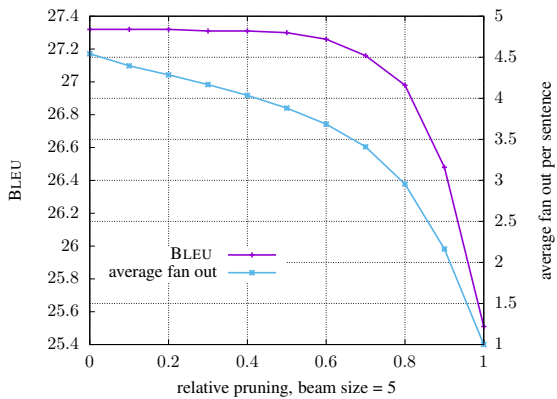}}
\vspace{-2em}
\caption{German$\rightarrow$English: Different values of relative pruning measured on newstest2014.}
\label{fig:rePruning}
\end{figure}

In Figure~\ref{fig:continue_ensemble}, you can see the German$\to$English translation performance and the average fan out per sentence for different beam sizes. Based on this experiment, we decided to run our pruning experiments for beam size 5 and 14.
The German$\to$English results can be found in Table~\ref{tab:results-deen}. 
By using the combination of all pruning techniques, we can speed up the decoding process by 13\% for beam size 5 and by 43\% for beam size 14 without any drop in performance. 
The relative pruning technique is the best working one for beam size 5 whereas the absolute pruning technique works best for a beam size 14. In Figure~\ref{fig:rePruning}
the decoding speed with different relative pruning threshold for beam size 5 are illustrated. Setting the threshold higher than 0.6 hurts the translation performance.
A nice side effect is that it has become possible to decode without any fix beam size when we apply pruning. Nevertheless, the decoding speed drops while the translation performance did not change. 
Further, we looked at the number of search errors introduced by our pruning schemes (number of times we prune the best scoring hypothesis). 
5\% of the sentences change due to search errors for beam size 5 and 9\% of the sentences change for beam size 14 when using all four pruning techniques together.

\begin{table*}[t!]
\begin{center}
    {\setlength{\tabcolsep}{.3em}
    \begin{tabular}{|l|c|c|c|c|cc|cc|}
        \hline
        \bf{pruning}& \bf{beam} & \bf{speed} & \bf{avg fan out} & \bf{tot fan out} & \multicolumn{2}{c}{\bf{MT08 nw}} & \multicolumn{2}{c|}{\bf{MT08 wb}} \\
                    & \bf{size} & \bf{up} & \bf{per sent} & \bf{per sent} & \BLEU & \TER  & \BLEU & \TER \\ \hline \hline
        no pruning & 1 & - & 1.00 & 29 & \rdm{27.34} & \rdm{61.66} & \rdm{25.97} & \rdm{60.32} \\ \hline
        no pruning & 5 & - & 4.36 & 137 & \rdm{34.43} & \rdm{57.33} & \rdm{30.64} & \rdm{58.23} \\
        rp=0.2 & 5 & 1\% & 4.32 & 134 & \rdm{34.43} & \rdm{57.33} & \rdm{30.57} & \rdm{58.21} \\ 
        ap=5 & 5 & 4\% & 4.26 & 132 & \rdm{34.32} & \rdm{57.27} & \rdm{30.63} & \rdm{58.19}\\ 
        rpl=0.01 & 5 & 1\% & 4.35 & 135 & \rdm{34.35} & \rdm{57.5} & \rdm{30.63} & \rdm{58.34} \\ 
        mc=3 & 5 & 0\% & 4.37 & 139 & \rdm{34.43} & \rdm{57.35} & \rdm{30.67} & \rdm{58.24} \\ 
        rp=0.2,ap=5,rpl=0.01,mc=3 & 5 & 10\% & 3.92 & 121 & \rdm{34.33} & \rdm{57.29} & \rdm{30.56} & \rdm{58.16} \\ \hline
        no pruning &  14 & - & 11.96 & 376 & \rdm{35.29} & \rdm{57.14} & \rdm{31.17} & \rdm{57.80} \\
        rp=0.2 & 14 & 3\% & 11.62 & 362 & \rdm{35.23} & \rdm{57.16} & \rdm{31.19} & \rdm{57.77} \\ 
        ap=2.5 & 14 & 14\% & 10.15 & 321 & \rdm{35.18} & \rdm{56.93} & \rdm{31.11} & \rdm{57.85} \\
        rpl=0.3 & 14 & 10\% & 10.93 & 334 & \rdm{35.27} & \rdm{57.24} & \rdm{31.06} & \rdm{57.89} \\
        mc=3 & 14 & 0\% & 11.98 & 378 & \rdm{35.34} & \rdm{56.92} & \rdm{31.12} & \rdm{57.82} \\
        rp=0.2,ap=2.5,rpl=0.3,mc=3 & 14 & 24\% & 8.62 & 306 & \rdm{35.29} & \rdm{56.86} & \rdm{31.08} & \rdm{57.76} \\ \hline
        rp=0.2,ap=2.5,rpl=0.3,mc=3 & - & - & 38.76 & 1411 & \rdm{35.24} & \rdm{57.29}& \rdm{31.06} & \rdm{57.94} \\ \hline
\end{tabular}
\caption{Results Chinese$\rightarrow$English: relative pruning(rp), absolute pruning(ap), relative local pruning(rpl) and maximum candidates per node(mc).}
\label{tab:results-zhen}
}
\end{center}
\end{table*}

The Chinese$\rightarrow$English translation results can be found in Table~\ref{tab:results-zhen}. 
We can speed up the decoding process by 10\% for beam size 5 and by 24\% for beam size 14 without loss in translation quality. 
In addition, we measured the number of search errors introduced by pruning the search. 
Only 4\% of the sentences change for beam size 5, whereas 22\% of the sentences change for beam size 14.

\section{Conclusion}
\label{sec:conclusion}
The original beam search decoder used in Neural Machine Translation is very simple. It generated translations from left-to-right
while looking at a fix number (beam) of candidates from the last time step only. By setting the beam size large enough, we ensure 
that the best translation performance can be reached with the drawback that many candidates whose scores are far away from the best are also explored.
In this paper, we introduced several pruning techniques which prune candidates whose scores are far away from the best one. By applying a combination of
absolute and relative pruning schemes, we speed up the decoder by up to 43\% without losing any translation quality.
Putting more diversity into the decoder did not improve the translation quality.

\bibliography{beamSearch}
\bibliographystyle{acl_natbib.bst}

\end{document}